\documentclass{article}

\usepackage[final,nonatbib]{nips_2016}

\pdfoutput=1
\usepackage[utf8]{inputenc} 
\usepackage[T1]{fontenc}    
\usepackage{url}            
\usepackage{booktabs} 
\usepackage{amsfonts}       
\usepackage{nicefrac}       
\usepackage{subfig}
\usepackage{microtype} 
\usepackage{times}
\usepackage{listings}
\usepackage{latexsym}
\usepackage{graphicx}
\usepackage{comment,bm}
\usepackage{color,amsmath}
\usepackage[square,numbers]{natbib}
\title{Stacking With Auxiliary Features}

\author{Nazneen Fatema Rajani\\
	   Dept. Of Computer Science\\
	    University of Texas at Austin\\
	    {\tt nrajani@cs.utexas.edu}
	  \And
	Raymond J. Mooney\\
  	 Dept. Of Computer Science\\
	    University of Texas at Austin\\
	    {\tt mooney@cs.utexas.edu}}

\begin{document}

\maketitle

\begin{abstract}
Ensembling methods are well known for improving prediction accuracy. However, they are limited in the sense that they cannot discriminate among component models effectively. In this paper, we propose stacking with auxiliary features that learns to fuse relevant information from multiple systems to improve performance. Auxiliary features enable the stacker to rely on systems that not just agree on an output but also the provenance of the output. We demonstrate our approach on three very different and difficult problems -- the Cold Start Slot Filling, the Tri-lingual Entity Discovery and Linking and the ImageNet object detection tasks. We obtain new state-of-the-art results on the first two tasks and substantial improvements on the detection task, thus verifying the power and generality of our approach.
\end{abstract}

\section{Introduction}
Using {\it ensembles} of multiple systems is a standard approach to improving
accuracy in machine learning \citep{dietterich:bkchapter00}. Ensembles have been
applied to a wide variety of problems in all domains of artificial intelligence including natural language processing and computer vision. However, these techniques do not learn to discriminate across the component systems and thus are unable to leverage them for improving performance. Thus combining these systems intelligently is crucial for improving the overall performance. We seek to integrate knowledge from multiple sources for improving ensembles of systems using Stacking with Auxiliary Features (SWAF). Stacking \citep{wolpert:nn92} uses supervised learning to train a meta-classifier to combine multiple system outputs. The auxiliary features enable the stacker to fuse additional relevant knowledge from multiple systems and thus leverage them to improve prediction accuracy.

\begin{figure*}
\centering
\includegraphics[height=6.6cm,width=10cm]{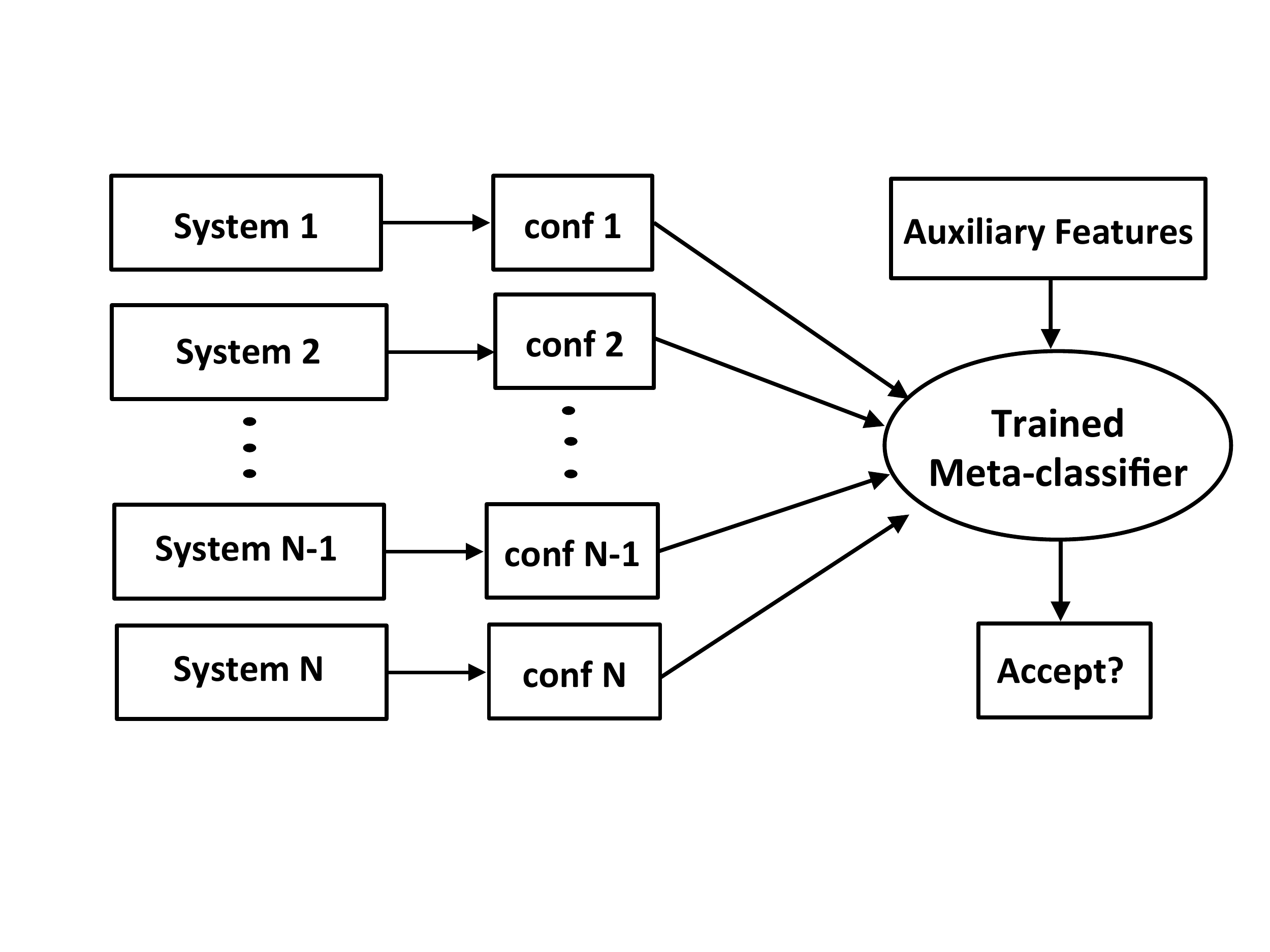}
\caption{Our stacking approach to combining system outputs using confidence scores and provenance as auxiliary features for improving prediction accuracy.}
\label{fig:sup}
\end{figure*}

In this paper, we consider the general problem of combining output from multiple systems to improve accuracy by using auxiliary features. Stacking with auxiliary features can be successfully deployed to any problem whose output instances have confidence scores along with {\it provenance} that justifies the output. Provenance indicates the origin of the generated output and thus can be used to measure reliability of system output. Figure~\ref{fig:sup} gives a generalized overview of our approach to combining multiple system outputs. The idea behind using auxiliary features is that a output is more reliable if not just multiple systems produce it but also agree on its provenance. For the CSSF task, provenance is the passage within the document from where the slot fill is extracted, for TEDL, it is the passage within the document that has the entity mention and for the object detection task, the bounding boxes serve as provenance for the detected object classes.

We use SWAF to demonstrate new state-of-the-art results for ensembling on three separate and unrelated tasks. The first two tasks are in the natural language processing domain and part of the NIST Knowledge Base Population (KBP) challenge -- {\it Cold Start Slot-Filling} (CSSF)\footnote{http://www.nist.gov/tac/2015/KBP/ColdStart/guidelines.html} and the {\it Tri-lingual Entity Discovery and Linking} (TEDL) \citep{ji:tac15}. The third task is in the domain of computer vision and part of the ImageNet 2015 challenge\citep{ilsvrc15} -- {\it Object Detection} from images. Our approach on these tasks outperforms the individual component systems as well as other ensembling methods such as the ``oracle'' voting baseline on all three tasks in the most recent 2015 competition; verifying the generality and power of stacking with auxiliary features for ensembling. 

\section{Background}
For the past several years, NIST has conducted
the English Slot Filling (ESF) and the Entity Discovery and Linking (EDL) tasks in the Knowledge Base Population (KBP) track as a part of the Text Analysis Conference (TAC). In 2015, the ESF task \citep{surdeanu:tac13,surdeanu:tac14} was replaced by the Cold Start Slot Filling (CSSF) task\footnote{http://www.nist.gov/tac/2015/KBP/ColdStart/index.html} which requires filling specific slots of information for a given set of query entities  based on a supplied text corpus. For the 2015 EDL task, two new foreign languages were introduced -- Spanish and Chinese as well as English -- and thus the task was renamed  Tri-lingual Entity Discovery and Linking (TEDL) \citep{ji:tac15}. The goal was to discover entities for all three languages based on a supplied text corpus as well as link these entities to an existing English Knowledge Base (KB) or cluster the mention with a NIL ID if the system could not successfully link the mention to any entity in the KB. The ImageNet object detection task is a widely known annual challenge for evaluating vision systems on a large real world corpus. The objective of the task is to produce produce
a list of object categories present in the image along
with an axis-aligned bounding box indicating the
position and scale of every instance of each object
category.

For CSSF, the participating systems
employ a variety of techniques such as such as relevant document extraction, relation-modeling, open-IE and inference. The top performing 2015 CSSF system \citep{angeli:tac15} leverages both distant
supervision \citep{mintz:acl09} and pattern-based relation extraction. Another system, {\it UMass\_IESL} \citep{roth:tac15}, used distant supervision, rule-based extractors, and semisupervised matrix embedding methods. The top performing 2015 TEDL system used a combination of deep neural networks and CRFs for mention detection and a language-independent probabilistic disambiguation model for entity linking  \citep{sil:tac15}. 

For the ImageNet object detection task in 2015, the top performing team used deep residual net \citep{he:arxiv15} and several other teams deployed a version of faster R-CNN with selective search \citep{renNIPS15fasterrcnn}. The faster R-CNN (Region based Convolutional Neural Networks) is a more efficient variant of fast R-CNN that uses Region Proposal Networks (RPN) to train an end-to-end network for generating region proposals. These proposals are then used by the fast R-CNN for object detection. 

In stacking, a meta-classifier is learned from the output of multiple
underlying systems. The stacker learns a classification boundary based on the confidence scores provided by the systems for each output instance. However, many times the scores produced by systems are not probabilistic or calibrated and cannot be compared meaningfully. In such circumstances, it is beneficial to also have other reliable auxiliary features like our approach. In the past, it has been shown that stacking multiple diverse systems improves performance on slot filling \citep{viswanathan:acl15}. However, our auxiliary features approach beats their ensemble resulting in a new state-of-the-art result on slot filling. There has been no past work on ensembling for the TEDL task and our approach beats the current best-state-of-the-art system. There has been some work on using stacked generalization to perform multi-layer object recognition \citep{peppoloni:ieee14} but our paper is the first to use stacking for ensembling multiple object detectors and we obtain substantial improvements over the component systems.

\section{Overview of the Tasks}
In this section we give a short overview of each of the three tasks considered in this paper to demonstrate successful deployment of our algorithm.

\subsection{Cold Start Slot Filling}
The goal of CSSF is to collect information (fills) about specific attributes (slots) for a set of entities (queries) from a given corpus. The queries entities can be a person (PER), organization (ORG) or geo-political entity (GPE). The slots are fixed and the 2015 task also included the inverse of each slot, for example the slot org:subsidiaries and its inverse org:parents. Some slots (like per:age) are
{\it single-valued} while others (like per:children) are {\it list-valued} i.e., they can take multiple slot fillers. 

The input for  CSSF is a set of queries and the corpus in which to look for information. The queries are provided in an XML format that includes an ID for the query, the name of the entity, and the type of entity (PER, ORG or GPE). The corpus consists of documents in XML format from discussion forums, newswire and the Internet,  each identified by a unique ID. The output is a set of slot fills for each query. Along with the slot-fills, systems must also provide its {\it provenance} in the corpus in the form {\it docid:startoffset-endoffset}, where {\it docid} specifies a source document and the offsets demarcate the text in this document containing the
extracted filler. Systems also provide a confidence score to indicate their certainty in the extracted information.

\subsection{Tri-lingual Entity Discovery and Linking}
The goal of TEDL is to discover all entity mentions in a corpus with English, Spanish and Chinese documents. The entities can be a person (PER),  organization (ORG), geo-political entity (GPE),  facility (FAC), or location (LOC). The FAC and LOC entity types were newly introduced in 2015. The extracted mentions are then linked to an existing English KB entity using its ID. If there is no KB entry for an entity, systems are expected to cluster all the mentions for that entity using a NIL ID.  

The input is a corpus of documents in the three languages and an English KB (FreeBase) of entities, each with a name,  ID,  type, and several relation tuples that allow systems to disambiguate entities. The output is a set of extracted mentions,  each with a string, its provenance in the corpus, and a corresponding KB ID if the system could successfully link the mention, or else a mention cluster with a NIL ID. Systems can also provide a confidence score for each mention.

\subsection{Object Detection for Images}
The goal of the object detection task is to detect all instances of object categories (out of the $200$ predefined categories) present in the image and localize them by providing coordinates of the axis-aligned bounding boxes for each instance. The ImageNet dataset is organized according to the WordNet hierarchy and thus the object categories are WordNet {\it synsets}.

The object detection corpus is divided into training, validation and test sets. The training set consists of approximately $450$K images including both positive and negative instances, annotated with bounding boxes; the validation set consists of around $20$K images also annotated for all object categories and the test set has $50$K images. The output for the task is the image ID, the object category ($1$-$200$), a confidence score and the coordinates of the bounding box. In case of multiple instances in the same image, each instance is mentioned on a separate line. 

\section{Methodology}
This section describes our approach to stacking multiple systems using their confidence scores and other auxiliary features. Figure~\ref{fig:sup} shows an overview of our system which trains a final meta-classifier for combining multiple systems. The auxiliary features depend on the task into consideration as described in the Section~\ref{sec:auxiliary}. 

\subsection{Stacking}
Stacking is a popular ensembling methodology in machine learning \cite{wolpert:nn92} and has been very successful in many applications including the top performing systems in the Netflix competition
\cite{sill:2009}. The idea is to employ multiple learners and combine
their predictions by training a ``meta-classifier'' to weigh and
combine multiple models using their confidence scores as features.  By
training on a set of supervised data that is disjoint from that used
to train the individual models, it learns how to combine
their results into an improved ensemble model that performs better than each individual component system. In order to successfully use stacking, the output must be represented as a {\it key-value} pair. The meta-classifier makes a binary decision for each distinct output pair. Thus before deploying the algorithm on a task, it is crucial to identify the {\it key} in the task output which serves as a unique handle for ensembling systems as well as the {\it values} which are results for a key provided by systems. Note that there is only one instance of a key in the output while there could be multiple values for a key from component systems. The output of the ensembling system is similar to the output of an individual system, but it productively aggregates results from different systems. In a final {\it post-processing} step, the outputs that get classified as ``correct" by the classifier are kept while the others are removed from the output.

The first step towards using the stacker is to represent the output as {\it key-value} pair. For the CSSF task, the {\it key} for ensembling multiple systems is a query along with a slot type, for example, per:age of ``Barack Obama'' and the {\it value} is a computed {\it slot fill}. For list-valued slot types such as org:subsidiaries, the key instance is repeated in the output for each value. For the TEDL task, we define the key to be the {\it KB (or NIL) ID} and the value to be a {\it mention}, that is a specific reference to an entity in the text. For the ImageNet object detection task, we represent the image ID as the {\it key} for ensembling and the {\it value} is a detected object category. The next step is to represent the output pair instances consistently. For a particular {\it key-value} pair if a system produced it then it also provides a confidence score else we use a confidence score of zero i.e. the output instance is incorrect according to that system. The output is now ready to be fed into the stacker as shown in Figure~\ref{fig:sup}. Along with confidence scores, we also feed in auxiliary features, described in the next section, for each task that enable the classifier to discriminate across component systems effectively and thus make better decisions.

\subsection{Auxiliary Features}
\label{sec:auxiliary}
As discussed earlier, systems must provide evidence in the form of provenance for each generated output pair. We use them as part of our auxiliary features that go into the stacker along with the confidence scores as shown in top part of Figure~\ref{fig:sup}. The provenance indicates origin or the source for the generated output instance and thus  depends on the task under considerations. For the CSSF task, if a system successfully extracts a relation then it must provide a slot filler provenance indicating the location of the extracted slot fill in the corpus. Provenance is more explicit for the two KBP tasks than the object detection task. For the KBP tasks, it serves as output justification in the form of text. On the other hand, for the detection task, output justification is in the form of bounding boxes and thus serves the same purpose as provenance. For the TEDL task, if a system successfully links a mention to a KB ID then it must provide the mention provenance indicating the origin of the mention in the corpus. For both the CSSF and TEDL tasks, the provenance is in the form of \emph{docid} and \emph{startoffset-endoffset} that gives information about the document in the corpus and offset in the document. On the other hand, for the ImageNet object detection task, if a system successfully detects a target category then it must provide the object bounding box localizing the object in the image. The bounding box is in the form of $\langle x_{min}, y_{min}, x_{max}, y_{max}\rangle$. The bounding box for object detection is similar to provenance for the KBP tasks and can thus be used as auxiliary features for stacking.

The idea behind using provenance as auxiliary features is that a output is more reliable if not just multiple systems produce it but also agree on the source/provenance of the decision. In order to enable the stacker to leverage the auxiliary features for discriminating among systems, we develop features that measure provenance similarity {\it across} systems. The Jaccard similarity
coefficient is one such statistical measure of similarity between sets and is thus useful in measuring the degree of overlap between the provenance provided by systems. For the CSSF and TEDL tasks, the provenance offsets(PO) are used to capture similarity as follows. For a given {\it key}, if $N$ systems that generate a {\it value} have the same \emph{docid} for their document provenance, then the provenance offset (PO) score is calculated as the intersection of offset strings divided by its union. Thus  systems that generate a {\it value} from
\emph{different} documents for the same {\it key} have zero
overlap among offsets.
\[PO(x)=\frac{1}{|N|}\times\sum_{i\in N,i\neq x}^{}\frac{|\textsf{substring(i)}\cap \textsf{substring(x)}|}{|\textsf{substring(i)}\cup \textsf{substring(x)}|}\]
For the object detection task, the Jaccard coefficient is used to measure the overlap between bounding boxes across systems. For a given image ID, if $N$ systems detect the same object instance, then the bounding box overlap (BBO) score is calculated as the the intersection of the areas of bounding boxes, divided by their union:
\[BBO(x)=\frac{1}{|N|}\times\sum_{i\in N,i\neq x}^{}\frac{|\textsf{Area(i)}\cap \textsf{Area(x)}|}{|\textsf{Area(i)}\cup \textsf{Area(x)}|}\]
We note that for the CSSF task, two systems are said to have extracted the same slot fill for a {\it key} if the fills are exactly same, however for the TEDL task, two systems are said to have linked the same mention for a {\it key} if the mentions overlap to any extent and finally for the ImageNet task, two systems are said to have detected the same object instance for a {\it key} if the Intersection Over Union (IOU) of the areas of their bounding boxes is greater than $0.5$. If the output {\it values} don't meet this criteria for a given {\it key}, then they are considered to be two different values for the same key. 

For the two KBP tasks, we also use the \emph{docid}
information as auxiliary features. For a given {\it key}, if $N$ systems provide a {\it value} and a
maximum of $n$ of those systems give the same \emph{docid} in their
provenance, then the document provenance score for those $n$ systems is $n/N$. Similarly, other systems are given lower
scores based on the fraction of systems whose provenance document
agrees with theirs.  Since this provenance score is weighted by the
number of systems that refer to the same provenance, it measures the
reliability of a {\it value} based on the document from where it originated. We note that the use of provenance
as features does not require access to the large corpus
of documents or image and is thus very computationally inexpensive.

Additional auxiliary features that we use are the slot type (e.g. per:age) for the CSSF task, and for the TEDL, we use the entity type and for the ImageNet task, we use the object category as an additional feature. For the CSSF task, features related to the provenance of the fill, as discussed above have also been used in \cite{viswanathan:acl15}. However, our novel auxiliary features for both the KBP tasks, that require access to the source corpus, further boosts our performance and beats their best ensemble. The 2015 CSSF task had a much smaller corpus of shorter documents compared to the previous year's slot-filling corpus \cite{ellis:tac15,surdeanu:tac14}. Thus, the provenance feature of \cite{viswanathan:acl15} did not sufficiently capture the reliability of a slot fill based on where it was extracted. Our new auxiliary feature measures the similarity between the {\it key} document and the {\it value} document. For the CSSF task, the {\it key} is the query entity along with slot type and thus the {\it key} document is the query document provided to participants to disambiguate query entities that could potentially have the same name but refer to different entities. For the TEDL task, the {\it key} is the KB ID of an entity and thus the {\it key} document is the pseudo-document made up of that entity's KB description as well as several relations involving the entity that exist in the KB. The document that the CSSF and TEDL systems provide as provenance is the {\it value} document for both the tasks. So the auxiliary feature uses cosine similarity to compare the {\it key} and {\it value} documents represented as standard TF-IDF weighted vectors. For the ImageNet task, we use the object class label as an additional feature. Some systems do well only on a set of categories such as deformable objects. Using the class label enables the stacker to learn to discriminate across such systems.
\begin{table*}[!ht]
\centering
\begin{tabular}{| c | c | c | c | }
\hline
 \rule{0pt}{2.5ex} 
\textbf{Methodology} & \textbf{Precision} & \textbf{Recall} & \textbf{F1} \\
 \hline
 \hline
 \rule{0pt}{2ex} 
Stacking with auxiliary features&0.4656&\textbf{0.3312}&\textbf{0.3871}\\ 
Stacking approach described in \citet{viswanathan:acl15}& \textbf{0.5084}&0.2855&0.3657\\
Top ranked CSSF system in 2015 \citet{angeli:tac15}&  0.3989&0.3058&0.3462\\
Oracle Voting baseline (3 or more systems must agree)&0.4384&0.2720&0.3357\\
 \hline
\end{tabular}
\caption{Results on 2015 Cold Start Slot Filling (CSSF) task using the official NIST scorer}
\label{table:results1}
\end{table*}
\subsection{Post-processing}
Once we obtain the decisions on each of the key-value pairs from the stacker, we perform some final post-processing so that the output from the stacker is as though it is generated by a single system. For CSSF, this is straight forward. Each list-valued slot fill that is classified as correct is included in the final output. For single-valued slot fills, if they are multiple fills that were classified correctly for the same query and slot type, we include the fill with the highest meta-classifier confidence. For TEDL, for each entity mention link that is classified as correct, if the link is a KB cluster ID then we include it in the final output, but if the link is a NIL cluster ID then we keep it aside until all mention links are processed. Thereafter, we resolve the NIL IDs across systems since NIL ID's for each system are unique. We merge NIL clusters across systems into one if there is at least one common entity mention among them. Finally, we give a new NIL ID for these newly merged clusters. 

For the ImageNet object detection task, for each object instance that is classified as correct by the stacker is included in the final output and the bounding box for that output instance is calculated as follows. If multiple systems successfully detected an object instance, then we sum the overlapping areas between a system's bounding box and every other system's that also detected the exact same instance and we do this for every such system. The bounding box produced by the system that has maximum overlapping area is included in the final output. Note that in case of two systems, this method is redundant and we include the bounding box produced by the systems with a higher confidence score.

\begin{table*}[!ht]
\centering
\begin{tabular}{| c | c | c | c | }
\hline
 \rule{0pt}{2.5ex}
\textbf{Methodology} & \textbf{Precision} & \textbf{Recall} & \textbf{F1} \\
 \hline
 \hline
  \rule{0pt}{2ex}
 Stacking with auxiliary features&0.803&\textbf{0.525}&\textbf{0.635} \\
Stacking approach described in \citet{viswanathan:acl15}& \textbf{0.814}&0.508  &0.625\\ 
Top ranked TEDL system in 2015 \citet{sil:tac15}&  0.693&0.547 &0.611\\
Oracle Voting baseline (4 or more systems must agree)&0.514 &0.601 &0.554\\ \hline
\end{tabular}
\caption{Results on 2015 Tri-lingual Entity Discovery and Linking (TEDL) task using the official NIST scorer and the CEAFm metric}
\label{table:results2}
\end{table*}

\begin{table*}[!ht]
\centering
\begin{tabular}[h]{| c | c | c | }
\hline
 \rule{0pt}{2.5ex}
\textbf{Methodology} & \textbf{Median AP} & \textbf{Mean AP} \\
 \hline
 \hline
 \rule{0pt}{2ex}
Stacking with auxiliary features&\textbf{0.526}&\textbf{0.546} \\
Best standalone system (VGG + selective search)\citet{renNIPS15fasterrcnn}&0.450&0.454\\
Oracle Voting baseline (1 or more systems must agree) &0.367&0.353\\
\hline

\end{tabular}
\caption{Results on 2015 ImageNet object detection task using the official ImageNet scorer.}
\label{table:results3}
\end{table*}

\section{Experimental Results}
\label{sec:results}

This section describes a comprehensive set of experiments evaluating SWAF on both the KBP tasks and the ImageNet object detection task using the algorithm described in the previous section, comparing our full system to various ablations and prior results. All KBP results were obtained using the official NIST scorers for the tasks provided after the competition ended.\footnote{http://www.nist.gov/tac/2015/KBP/ColdStart/tools.htm, https://github.com/wikilinks/neleval}. For the object detection task, the results are obtained using the scorer provided with ImageNet devkit.  

SWAF relies on training data for learning and thus for the two KBP tasks, we only use systems that participated in both 2014 and 2015 iterations of the tasks. This allows us to train the stacker on 2014 system outputs and use the trained model to evaluate on the 2015 iteration of the tasks. In this way we used $10$ common systems for the CSSF task and $6$ for the TEDL task. We were unable to obtain the system outputs directly for any iteration of the ImageNet task and so we use two pre-trained deep neural models on the ImageNet object detection training set, the ZF and the VGG models described in \citep{renNIPS15fasterrcnn}. We run these models on the validation set using the faster-RCNN method\citep{renNIPS15fasterrcnn} with selective search\citep{uijlings2013selective} on top of Caffe\citep{jia2014caffe}. We also use the Deformable Parts Model (DPM) \citep{felzenszwalb2010object} with selective search for object detection. The DPM models takes really long to process each image and was unable to process the entire test set in time on all $200$ categories. Therefore, we run all our experiments only on the validation set because based on the competition policies, we are heavily penalized for submitting partial output on the test set. We divide the validation set into three equal parts and train on two thirds of the set and test on the remaining one third set. We had a total of $3$ systems as a part of the ensemble for the object detection task.

For the CSSF task, systems are evaluated against a gold standard using precision, recall and F1 scores based on the slot fills that a system could successfully extract.  The TEDL evaluation provides three different approaches to measuring Precision, Recall and F1. First is entity discovery, second is entity linking and last is mention CEAF \citep{ji:tac15}. The mention CEAF metric finds the optimal alignment between system and gold standard clusters, and then evaluates precision and recall micro-averaged
over mentions. We obtained similar results on all three evaluations and thus only include the mention CEAF score in this paper. For the ImageNet challenge, the detection task is judged by the
average precision (AP) on a precision/recall curve and a predicted
bounding box of a class is considered correct if its intersection over union with the ground truth exceeds a threshold of $0.5$\citep{ilsvrc15}. The output from the scorer is AP for each of the $200$ classes along with the median AP and and mean AP. We only report the median AP and mean AP (mAP) in this paper.

We compare our results to several baselines. For the two KBP tasks, we compare against the stacking approach of \citep{viswanathan:acl15} by evaluating on systems that are common between 2014 and 2015. The authors of \citep{viswanathan:acl15} only report results on the CSSF task and we run their system on the TEDL task for the sake of comparison. We also compare to the top ranked systems for both the CSSF and TEDL tasks in 2015 as well as the voting baseline for ensembling the system outputs. For this approach, we vary the
threshold on the number of systems that must agree to identify an ``oracle'' threshold that
results in the highest F1 score for 2015 by plotting a Precision-Recall curve and finding the best F1 score for the voting baseline for each of the three tasks. At each step we add one more to the number of systems that must agree on a key-value. We find that for CSSF, a threshold of $3$ or more systems and for TEDL a threshold of $4$ or more systems gives us the best resulting F1 for voting. For the object detection task, a threshold of $1$ (i.e. the union of all systems) gives us the best resulting mAP. We note that oracle voting is ``cheating'' to improve the standard voting baseline. 

\begin{figure}[!ht]
\captionsetup[subfigure]{labelformat=empty}
\centering
\includegraphics[width=.5\textwidth,height=3cm]{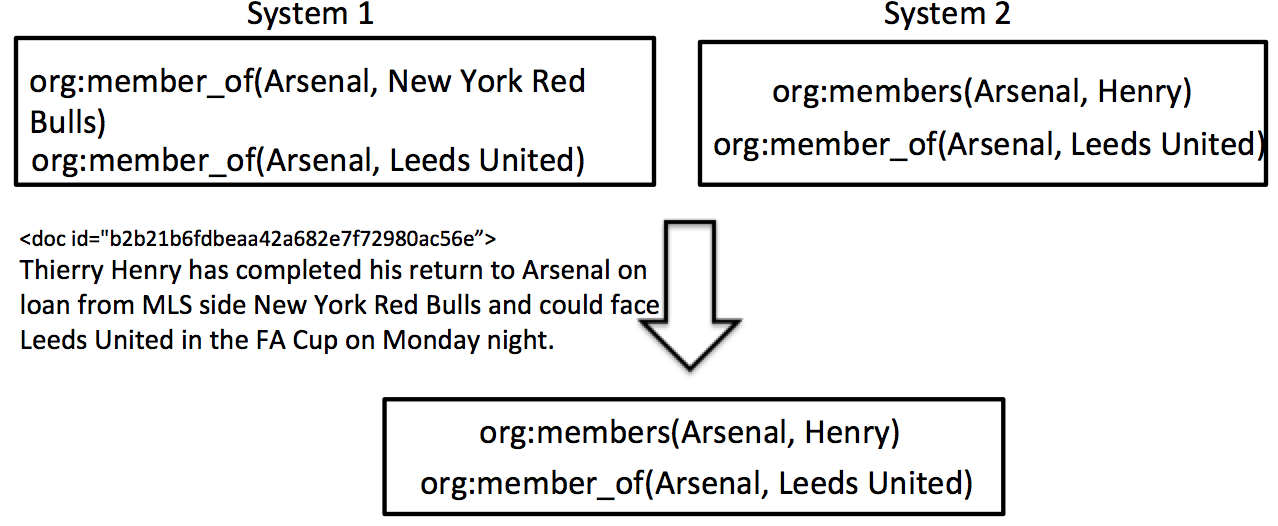}\hfill
\includegraphics[width=.5\textwidth,height=3cm]{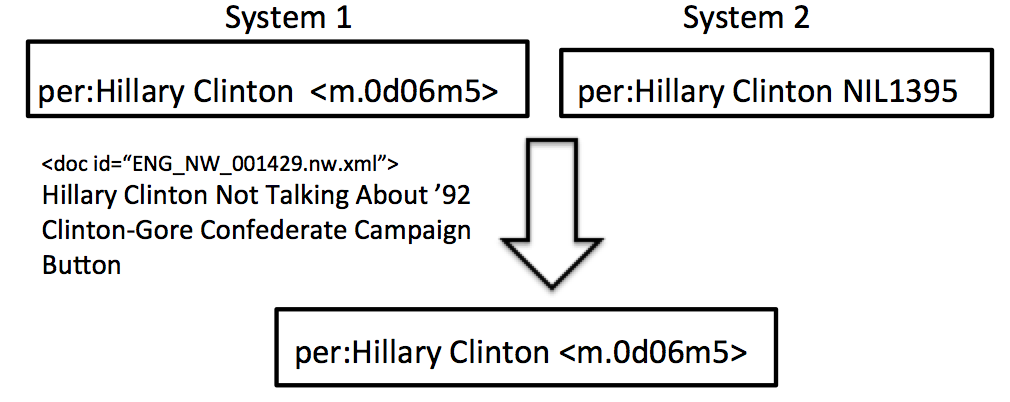}
\subfloat[]{\includegraphics[width=.5\textwidth,scale=0.5,height=6cm]{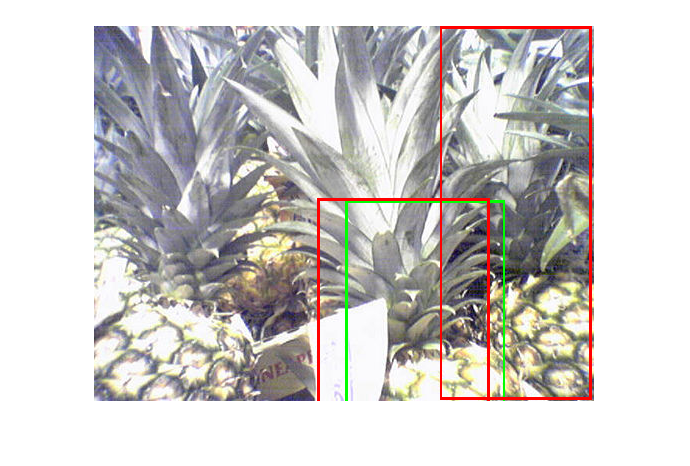}}
\subfloat[]{\includegraphics[scale=0.5,height=6cm,width=.5\textwidth]{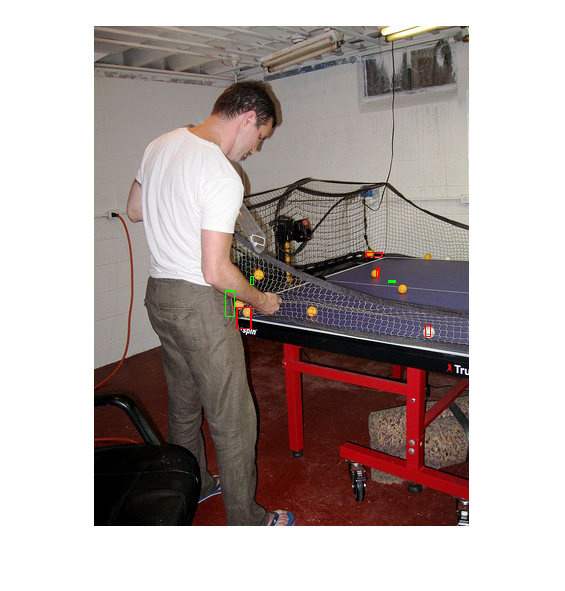}}
\vspace{-0.2cm}

\caption{Sample outputs obtained using SWAF on various 2015 tasks. The top left shows output from the CSSF KBP task while on the top right is the TEDL task. The snippet next to the arrow displays the provenance used by SWAF while classifying. The bottom row is output obtained on the ImageNet object detection task. The green bounding box are those obtained by the systems and among those, the red ones are classified correct by SWAF.}
\label{fig:results}
\end{figure}

Tables \ref{table:results1} and \ref{table:results2} show the results for CSSF and TEDL respectively. SWAF performed consistently on both the tasks beating all baseline ensembles as well as the top ranked systems for 2015.  The oracle voting baseline performs very poorly indicating that naive ensembling is not advantageous. The relative ranking of the approaches is similar to those obtained on both the CSSF and TEDL tasks, thus proving that our approach is very general and provides improved performance on two quite different and challenging KBP problems. The top part of Figure~\ref{fig:results} shows sample output obtained from the two KBP tasks by using SWAF. The left part shows the CSSF task, two systems produce one common relation and one different relation along with provenance justifying the document where the relation is extracted. This enables SWAF to better decide on the correct relations. Similarly the right side shows the TEDL task sample output. Although many systems link the entity mention of ``Hillary Clinton'' in the document to NIL, SWAF is able to make the right decision based on the provenance and links it to the correct cluster ID in FreeBase.

Table~\ref{table:results3} shows the results obtained on the ImageNet 2015 object detection task. SWAF beats the best standalone system as well as the oracle voting baseline by a large margin. For the voting baseline, we consider an object instance to be the same while voting, if its bounding boxes produced by the systems have IOU greater than $0.5$. We found that we get the best voting baseline if we just take the union of all outputs produced by systems. On analyzing the results, we found that the AP of several object classes differed widely across systems and even more so between the deep systems and DPM. Using SWAF, the classifier learns to discriminate systems based on the auxiliary features and is thus able to leverage component systems in the overall performance. The bottom half of Figure~\ref{fig:results} shows sample of results obtained by SWAF on two images. We chose these images because the component systems had high variance in performance on these two classes -- ``ping pong ball (class 127)'' and ``pineapple (class 126)''. The  figure shows bounding boxes obtained by component systems for only these two categories respectively. Although the image is very cluttered, SWAF uses the output bounding boxes as context to make better classification decisions. Based on the outputs obtained from SWAF, we found that our approach does well on localizing objects in images that have multiple instances of the same object, i.e. the image can be considered to be ``cluttered''.

\vspace{-0.5cm}
\section{Conclusion}
In this paper, we present stacking with auxiliary features, a novel approach to ensemble multiple diverse system outputs. The auxiliary features enable information fusion and thus can be used to discriminate among component systems. We demonstrate that our approach can be generalized by applying it on three very different tasks, the Cold Start Slot Filling and the Tri-lingual Entity Discovery and Linking tasks in the NLP domain and the ImageNet object detection task in computer vision. We obtain very promising results on all three tasks, beating the best component systems as well as other baseline ensembling methods. The approach provides an overall F1 score of $38.7\%$ on
2015 KBP CSSF task and CEAFm F1 of $63.5\%$ on 2015 KBP TEDL, and an overall mAP of $54.6\%$ on the ImageNet object detection task. We achieve a new state-of-the-art on the two KBP tasks and substantial improvements over baselines on the detection task.

On analyzing the results obtained, we find that SWAF does better when the component systems differ widely on their outputs and have low confidences. This leads us to conclude that the gain in performance from SWAF comes from output decisions that are difficult to make without context but using auxiliary features enables fusion of additional relevant information, allowing the stacker to make the right decision.
\section{Acknowledgment}
This research was supported in part by the DARPA DEFT program under AFRL grant FA8750-13-2-0026 and by MURI ARO grant W911NF-08-1-0242.
\bibliography{nips2016}
\end{document}